\documentclass[11pt]{article}
\usepackage{acl}
\usepackage{dblfloatfix}
\usepackage{amsmath, amssymb}
\usepackage{graphicx}
\usepackage{hyperref}
\usepackage{geometry}        
\usepackage{times}           
\usepackage{float}           
\usepackage{booktabs}        
\usepackage{caption}
\usepackage{array}
\usepackage{subcaption}
\usepackage{multirow}
\usepackage{relsize} 
\usepackage{placeins}
\usepackage{microtype}
\usepackage{xurl}

\title{\Large Efficiently Adapting Spoken Language Models for the Singaporean Context}
\author{
  \large \textbf{Language AI R\&D, xData} \\
  \normalsize Home Team Science \& Technology Agency (HTX), Singapore \\
  \\
  Ng Jia Sheng Jason \\
}
\date{July 10, 2026}

\begin{document}

\maketitle

\begin{abstract}
Spoken language models (SLMs) unify speech perception and reasoning, but adapting them to sensitive domains is underexplored, especially when the original training data is inaccessible and the use case demands multilingual, spoken-query interaction. We adapt an open-source SLM to the Singaporean Home Team context across five speech tasks in Singapore's four official languages, combining LoRA fine-tuning, a surrogate text-QA dataset that guards against catastrophic forgetting, and a multi-task objective that adapts the CoBa reweighting scheme to speech. We also build HTD-multilingual-QA, a 504,853 sample multilingual QA dataset in text and spoken form. The resulting HT-Moonstone (5B) matches or outperforms SLMs up to 7× its size on most tasks, attains the best accent and gender recognition among all models evaluated, and loses under 2\% of its original speech QA ability.
\end{abstract}



\section{Introduction}
The Home Team Science and Technology Agency (HTX) is a statutory board under Singapore's Ministry of Home Affairs (MHA). It serves as a ``force multiplier'' for the Home Team departments, applying science and technology to strengthen homeland security and public safety. Among the use cases HTX receives from these departments are speech tasks such as automatic speech recognition (ASR) and spoken question answering (QA), the latter forming the core of speech-to-speech conversational AI applications. Because much of this work involves highly sensitive data,  closed-source or proprietary models are often unsuitable, which is why we turn to open-source alternatives.\\

HTX's Language AI R\&D team has been developing and adapting open-source foundational speech models, such as those for ASR. However, the field is moving towards spoken language models (SLMs) \citep{slm_survey}, which handle a wide range of cognitive and perception speech tasks \citep{slm_tasks}. Several open-source SLMs already exist, including Kimi-Audio \citep{kimiaudio} and Audio-Flamingo-Next \citep{audioflamingo}, as well as models built for the Singaporean context, such as MERaLiON-2 \citep{meralion}. However, because our operating environment and use cases are specific to the Home Team context, a suitable model must either be built from scratch or adapted from an existing open-source SLM. Training from scratch demands far more audio data, which is costly to acquire, so we favour adaptation. The SLM domain is still nascent, and few studies document how to adapt these models effectively for new tasks while preserving their original performance, particularly when the full dataset used to build them is unavailable.\\

As we have use cases involving the use of text context and spoken queries and no Singaporean multilingual spoken QA training dataset currently exists, the development of a resource that covers Singaporean English, Mandarin, Bahasa Melayu, and Tamil is essential for adapting models to handle such spoken QA task effectively. Furthermore, adapting models without full access to their original training data often risks catastrophic forgetting. Our adaptation strategy is designed to mitigate this risk, ensuring we maximise performance on both perceptive and cognitive speech tasks in the Singaporean domain while preserving the model's foundational capabilities.\\

In this paper, we address these challenges and introduce:
\begin{enumerate}
    \item a strategy to effectively adapt SLMs to the Singaporean context for multiple speech tasks;
    \item \textbf{HTD-multilingual-QA}: a multilingual, multi-turn QA training dataset in both text and spoken form, grounded in Singaporean Home Team context that covers the 4 Singaporean languages - English, Mandarin, Bahasa Melayu, and Tamil for training SLMs on spoken QA tasks; and
    \item \textbf{HT-Moonstone}, an adapted 5B-parameter SLM that performs competitively for both perceptive and cognitive speech tasks in the Singaporean domain against open-source models up to 7$\times$ its size while exhibiting only a 2\% degradation in its original speech QA capabilities.
\end{enumerate}

\begin{figure*}[tp]
    \centering
    \includegraphics[width=\textwidth]{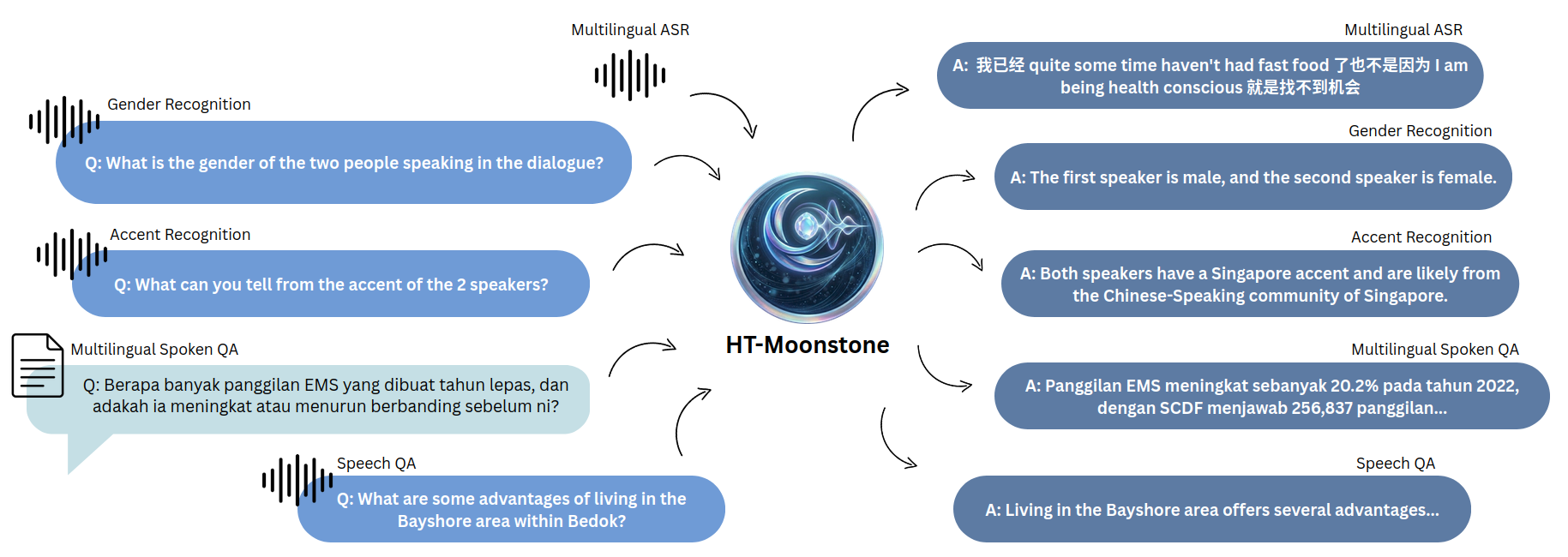}
    \caption{Examples of target tasks that can be handled by HT-Moonstone. Inputs for Multilingual Spoken QA are different (text context, audio query) compared to the rest of the tasks (audio context, text query).}
    \label{fig:ht-moonstone-tasks}
\end{figure*}

\section{Related Work}
\subsection{Rise of Spoken Language Models}
\label{sec:rise-slm}
The paradigm of speech processing is transitioning from specialised, single-task architectures to versatile, multi-task capable SLMs. Commonly known approaches typically unify audio encoders such as Whisper \citep{whisper} with pre-trained Large Language Models (LLMs) as decoders via an intermediate adapter layer. This architecture maps audio embeddings into a latent space compatible with textual tokenisation, enabling direct audio-to-text processing. By replacing conventional cascaded pipelines, which decouple ASR from downstream reasoning, these models not only reduce latency but also mitigate the \textit{"broken telephone"} effect, which is the propagation of transcription errors that inherently undermines performance in downstream speech QA tasks. Processing audio representations directly also preserves paralinguistic information beyond lexical content, enabling the decoder to reason over cues such as prosody and affect for tasks like emotion recognition. Examples of SLMs include Voxtral \citep{voxtral}, Audio Flamingo \citep{audioflamingo} and MERaLiON-2 \citep{meralion}, which has been developed specifically to address the nuances of the Singaporean context. \\

Despite the emergence of these models, adapting them to the needs of specific, sensitive environments such as the Singaporean Home Team remains an open challenge. Existing adaptation literature are often limited to single task adjustments such as for ASR. Furthermore, existing SLM frameworks are frequently developed for scenarios involving text-based queries, leaving the more complex paradigm of text-context and spoken-query interaction relatively underexplored.

\subsection{Singaporean Audio-Text Datasets}
The linguistic landscape of Singapore is defined by a quadrilingual standard of English, Mandarin, Bahasa Melayu, and Tamil, which is often characterised by unique grammatical structures and code-mixing. These features are exemplified by \textit{Singlish}, where localized syntax and lexicon are frequently interspersed with vocabulary from regional languages. For instance, the query "\textit{Eh, you makan already a not ah?}" (which means "Hey, have you eaten?") illustrates the seamless integration of the Bahasa Melayu term \textit{makan}, meaning eat, into a colloquial English syntactic frame complete with particle-driven modality. Such phenomena, where conversational intent is conveyed through a fluid combination of tone, particles, and multilingual lexicon, create a significant challenge for generic spoken language models trained on monolithic, standardised datasets. The specific prosody and conversational dynamics inherent to the unique Singaporean speech necessitate the development of localised datasets to effectively adapt open-source models to the Singaporean domain. While existing resources like the Multitask-National-Speech-Corpus (MNSC) \citep{mnsc} have provided a foundation for tasks like ASR and speech QA, defined therein as audio-based contexts paired with textual queries, they do not fully encapsulate the complexity of spontaneous, multilingual conversational speech. Crucially, our operational use cases to use SLMs for speech-2-speech conversational AI require a paradigm shift because the model must process textual context to ensure factual grounding while responding to spoken queries across all four Singaporean languages. While SQuAD-SRC \citep{squadsrc} encompasses the idea of spoken queries, it does not cover the spoken style of Singaporean speech nor the other 3 Mother Tongue languages spoken in Singapore. This establishes a clear research opportunity to develop such resources.

\subsection{Adapting SLMs}
Two broad strategies exist to adapt SLMs for a particular domain, language and/or tasks. The first is to assemble pre-trained encoder and decoders and align the whole architecture over large-scale corpora; this is the route taken by the Singapore-oriented MERaLiON-2 \citep{meralion} model, as well as by Southeast Asian systems such as Typhoon-Audio \citep{typhoonaudio} and SeaLLMs-Audio \citep{seallmsaudio}. While effective, this strategy is expensive and presupposes access to hundreds of thousands of audio hours of training data with huge computing power, making it ill-suited to data and resource-constrained conditions. \\

The second strategy, and the one we pursue, is the post-pretraining adaptation of an existing, pretrained SLM through parameter-efficient fine-tuning (PEFT). This route is considerably more economical, yet it remains comparatively underexplored. CantoASR \citep{cantoasr} instruction-tunes Qwen2-Audio with LoRA \citep{lora} for low-resource Cantonese tone and accent correction, while \citep{slu_limited_data} cross-lingually adapts Qwen2-Audio to low-resource languages from minimal target-language speech. Most relevant to the Singaporean context, \citep{dpo_codeswitch} applies direct preference optimisation to MERaLiON-2, Phi-4-multimodal, and Qwen2-Audio to improve English--Mandarin code-switched recognition on Singapore--Malaysia conversational speech. As foreshadowed in Section~\ref{sec:rise-slm}, however, these efforts share two limitations: they overwhelmingly optimise a single transcription objective (ASR), and they target the standard text-query setting rather than the spoken-query, text-response interaction central to one of our use cases. To our knowledge, no prior work adapts a pretrained SLM to both perception and cognitive speech tasks via a multi-task training approach, especially for the spoken-query setting. Our work addresses this gap, demonstrating that an existing SLM can be efficiently adapted to the Singaporean domain via LoRA \citep{lora}, without the cost of constructing a Singaporean model from scratch.

\section{SLM Adaptation Methodology}
\subsection{Base Model Selection}
\label{sec:base-model-selection}
\begin{table*}[tp]
    \centering
    \caption{Zero-shot benchmark of candidate base SLMs across the five target tasks, arranged in ascending order of model size. Details of datasets used for each tasks are listed in Appendix \ref{app:first}.
Best result per column in \textbf{bold}. ($\uparrow$/$\downarrow$ indicate that
higher/lower is better.)}
    \label{tab:base-model-benchmark}
    \begin{tabular}{lcccccc}
        \toprule
        & & \textbf{ASR - EN} & \textbf{AR} & \textbf{GR} & \textbf{Spoken QA} & \textbf{Speech QA} \\
        \textbf{Model} & \textbf{Size (B)} & WER\,$\downarrow$ & Acc.\,$\uparrow$ & Acc.\,$\uparrow$ & Acc.\,$\uparrow$ & Acc.\,$\uparrow$ \\
        \midrule
        MERaLiON-2-3B               & 3.5 &     \textbf{5.05} & 44.4 & \textbf{87.7} & 25.4 & 68.2 \\
        Voxtral-Mini-3B-2507        & 5 &       7.25 & 24.7 & 15.7 & 42.2 & 76.4 \\
        audio-flamingo-next-hf      & 8 &       7.87 & 0 & 74.6 & 33.0 & 68.7 \\
        SeaLLMs-Audio-7B            & 8 &       114 & 0 & 38.6 & 17.25 & 40.6 \\
        Kimi-Audio                  & 10 &      8.36 & 3.7 & 79.8 & 48.6 & 71.3 \\
        MERaLiON-2-10B              & 10 &      5.46 & \textbf{56} & 75.5 & 45.3 & 73.5 \\
        Qwen2.5-Omni                & 11 &      188 & 0 & 18.2 & 36.0 & 79.0 \\
        Voxtral-Small-24B-2507      & 24 &      7.64 & 8.3 & 0.5 & 42.5 & \textbf{80.0} \\
        Qwen3-Omni                  & 35 &      9.85 & 18.6 & 79.1 & \textbf{61.6} & 76.5 \\
        \bottomrule
    \end{tabular}
\end{table*}

Prior to any fine-tuning, we first establish which pretrained SLM offers the strongest foundation for adaptation to the Singaporean domain. To this end, we benchmarked a set of candidate open-source models across five perception and cognitive speech tasks chosen to reflect our target use cases:
\begin{enumerate}
  \item \textbf{Automatic Speech Recognition (ASR) - EN:} transcription of Singaporean-accented English.
  \item \textbf{Accent Recognition (AR):} identification of ethnic groups among Singaporean speakers.
  \item \textbf{Gender Recognition (GR):} identification of Singaporean speakers' gender.
  \item \textbf{Spoken Question Answering (Spoken QA):} generation of a textual answer, given a spoken question in one of the four official languages.
  \item \textbf{Speech Question Answering (Speech QA):} generation of a textual answer, given a Singaporean context in spoken form and a textual question in English.
\end{enumerate}

The evaluation sets for ASR, AR, and GR are drawn from the MNSC dataset (ASR-PART1-Test, PQA-AR-Dialogue-Test, and PQA-GR-Dialogue-Test, respectively). The Speech QA set comprises of textual questions using public Singaporean speech as context \citep{audiobench} while the Spoken QA set was developed in-house, as detailed in Section~\ref{sec:htd-multilingual-qa}. All models were evaluated using an adapted version of the AudioBench \citep{audiobench} evaluation framework with a GPT-4o judge and the default evaluation prompts. Table~\ref{tab:base-model-benchmark} reports each candidate's results. No single model dominated across all five tasks. We observed some outlier results; on investigation, these arose because the model either declined to answer or produced responses that were insufficiently specific. We selected \textbf{Voxtral-Mini-3B-2507} as our base model on the basis of three considerations: (1) its competitively low WER for Singaporean English ASR task, (2) its comparable Speech QA accuracy given its small size, and (3) its capacity to process up to 40 minutes of audio in a single pass \citep{voxtral}, which is essential for our long-form audio analysis use cases. While its accent and gender recognition accuracy is comparatively low, we prioritised speech QA performance, which more closely reflects our intended use cases.

\subsection{Training Datasets}
\begin{table*}[tp]
\centering
\caption{Maximum available training samples, grouped by task domain. ASR is further split into language sub-domains.}
\label{tab:train-dataset-sizes}
\setlength{\tabcolsep}{4pt}
\resizebox{\textwidth}{!}{%
\begin{tabular}{lllr}
\toprule
Task Domain & Sub-domain & Dataset & Max. Samples \\
\midrule
\multirow{10}{*}{ASR}
& English [EN]  & MNSC ASR (PART1,2,3,5,6)  & 4{,}956{,}791 \\
\cmidrule(l){2-4}
& \multirow{4}{*}{Mandarin [ZH]}
                & AISHELL-1                 & 120{,}098 \\
&               & SEAME (ZH-CS)         & 95{,}853 \\
&               & AISHELL-3                 & 56{,}936 \\
&               & FLEURS [ZH]               & 3{,}246 \\
\cmidrule(l){2-4}
& \multirow{2}{*}{Malay [MS]}
                & Mesolitica (MS-CS)        & 17{,}851 \\
&               & FLEURS [MS]               & 2{,}667 \\
\cmidrule(l){2-4}
& \multirow{3}{*}{Tamil [TA]}
                & SLR127                    & 69{,}583 \\
&               & SLR65                     & 3{,}427 \\
&               & FLEURS [TA]               & 2{,}367 \\
\midrule
Accent Recognition (AR)    & & MNSC PQA-AR & 4{,}862{,}485 \\
\midrule
Gender Recognition (GR)    & & MNSC PQA-GR & 4{,}862{,}485 \\
\midrule
Text Instruction Following & & Nemotron-SFT-IF-Chat-v2 (reasoning\_off)$^{\dagger}$ & 1{,}063{,}555 \\
\midrule
\multirow{2}{*}{Conversational QA}
& & HTD Multilingual QA (Text)   & 504{,}853 \\
& & HTD Multilingual QA (Spoken) & 504{,}853 \\
\midrule
\multicolumn{3}{l}{\textbf{Total}} & \textbf{17{,}127{,}050} \\
\bottomrule
\end{tabular}%
}
\\[4pt]
\footnotesize $^{\dagger}$Estimated from average line size; exact count not precomputed.
\end{table*}

Guided by the tasks of interest established in Section~\ref{sec:base-model-selection}, we curated the training datasets listed in Table~\ref{tab:train-dataset-sizes} to adapt the base SLM. The table reports the maximum number of samples available in each dataset, categorised by their dataset domains; the actual training mix is sub-sampled according to a fixed proportion rather than using all samples. The development of the HTD-Multilingual-QA datasets (in both text and spoken form) is described in Section~\ref{sec:htd-multilingual-qa}, and our motivation for including the Nemotron-SFT-Instruction-Following-Chat-v2 dataset is discussed in Section~\ref{sec:chat-grounding}.
  
\subsubsection{Development of HTD-multilingual-QA Dataset}
\label{sec:htd-multilingual-qa}
To develop conversational models capable of understanding Singaporean-style inputs, we previously built a Singaporean multi-turn conversational QA dataset grounded in Home Team context\footnote{\url{https://medium.com/htx-dsai/how-we-talk-generating-singaporean-conversations-54aea9b6892b}} and based it on a range of Singaporean personas. Building on this dataset, we augmented it by translating equal proportions of the user turns into the other Singaporean languages (code-mixed with English) using GPT-4o, yielding the textual variant of the HTD-Multilingual-QA dataset. The user turns were then converted to spoken form with OmniVoice \citep{omnivoice}, using voice cloning seeded from two manually recorded audio clips of Singaporean speakers per language. The recorded voices stay true to the speakers' ethnicities; for example, speakers of Chinese ethnicity were recorded speaking code-switched Mandarin. During spoken-form generation, we found that roughly 0.2\% of the translated user turns were erroneous, exhibiting repeated tokens or altered meaning; these were corrected afterwards. Assistant turns were not converted to spoken form. In total, this yields 504{,}853 samples comprising 1{,}723{,}693 QA pairs distributed across 2 to 8 turns, in each of the text and spoken variants.

\subsubsection{Usage of Surrogate Datasets}
\label{sec:chat-grounding}
Since we intend to adapt the SLM chosen in Section \ref{sec:base-model-selection} to other speech tasks in the Singaporean domain, we mitigate catastrophic forgetting through PEFT. Beyond this, we specifically aim to preserve the model's original instruction-following and QA capabilities. Because the original training data is unavailable to us, we instead seek to identify a surrogate dataset capable of maintaining the model's speech QA performance during adaptation. To this end, we curated a set of both text QA and speech QA datasets and evaluated their effectiveness in preserving spoken QA and speech QA capabilities in the English language. For our benchmarks, we focus on both speech QA and spoken QA rather than text QA capabilities because this is what our downstream use cases require.\\

\begin{table}[t]
\centering
\footnotesize
\setlength{\tabcolsep}{3.5pt}
\caption{Text/Speech QA candidate surrogate dataset benchmarks. Details of datasets used for each tasks are listed in Appendix \ref{app:first}.}
\label{tab:test-qa-dataset}
    \begin{tabular}{@{}>{\raggedright\arraybackslash}p{0.44\columnwidth}cc@{}}
        \toprule
        & \textbf{Spoken QA (EN)} & \textbf{Speech QA} \\
        \textbf{Condition/Dataset} & Acc.\,$\uparrow$ & Acc.\,$\uparrow$ \\
        \midrule
        Control (no-training)                           & 61.8 & 76.4 \\
        \midrule
        dolphin                                         & 64.3 & 76.0 \\
        OpenHermes-2.5                                  & 55.6 & 74.2 \\
        Nemotron-SFT-Instruction-Following-Chat-v2      & \textbf{64.8} & \textbf{79.7} \\
        no\_robots                                      & 52.4 & 69.5 \\
        ragbench                                        & 61.4 & 75.1 \\
        LongAudio (MultiDialog)                         & 53.8 & 66.8 \\
        MNSC (SQA)                                      & 45.0 & 62.8 \\
        \bottomrule
    \end{tabular}
\end{table}

Across our PEFT experiments on each candidate dataset listed in Table~\ref{tab:test-qa-dataset}, in which the SLM was fine-tuned on 20k samples each, Nemotron-SFT-Instruction-Following-Chat-v2 emerged as the most effective at mitigating catastrophic forgetting. It was the only dataset to improve performance on both tasks; the remaining datasets degraded one or both scores. Notably, it achieved these gains despite being a text-only QA dataset, whereas the evaluation operates directly on speech-modality inputs.

\subsection{Multi-Task Training}
\label{sec:multi-task-training}
Adapting the SLM to the Singaporean context introduces additional complexity when several task domains must be learned simultaneously, which leads to two identified issues. Firstly, the per-batch loss cannot simply be aggregated over all tokens; deeper cognitive speech task domains such as QA emit far more tokens than shallow cognitive task domains such as gender recognition, so token-level averaging systematically suppresses the contribution of the latter. Secondly, the evaluation loss of each task domain converges at a different point during training, so tasks that converge early are prone to overfitting while later-converging tasks are still improving. \\

To address the first issue, we decouple a sample's loss from its answer length by averaging the cross-entropy over each sample's own answer tokens before aggregating across samples,

\begin{equation}
\begin{aligned}
  \ell(x)       &= \frac{1}{|y|}\sum_{j=1}^{|y|}\mathrm{CE}\!\left(y_j, \hat{y}_j\right), \\
  \mathcal{L}_i &= \frac{1}{|B_i|}\sum_{x \in B_i}\ell(x).
\end{aligned}
\end{equation}

where $y$ is the target answer of sample $x$, $B_i$ is the set of samples in the batch belonging to task $i$, and $\mathcal{L}_i$ is the resulting per-task-domain loss. The per-task-domain loss is aggregated through the use of weights that can be adjusted throughout the training process,

\begin{equation}
    \mathcal{L}(t) = \sum_{i=1}^{K} \omega_i(t)\mathcal{L}_i .
\end{equation}

\begin{figure*}[tp]
    \centering
    \includegraphics[width=\textwidth]{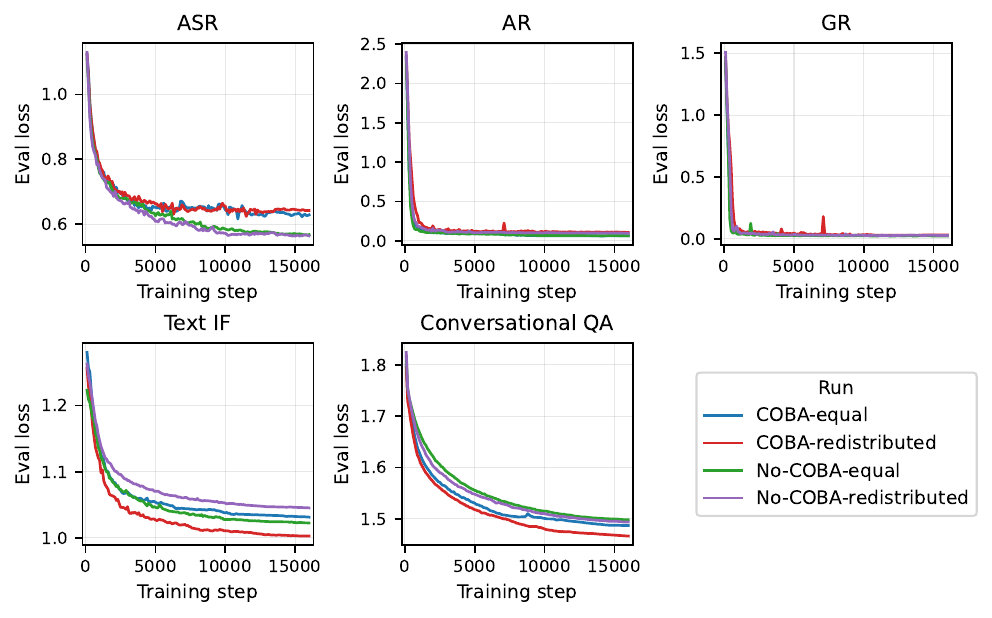}
    \caption{Per dataset domain evaluation loss over training for the four 256k-sample runs, under equal versus redistributed dataset proportion conditions and with or without CoBa.}
    \label{fig:eval-loss}
\end{figure*}

The second issue concerns task domains converging at different points, which leaves early-converging task domains prone to overfitting while later ones are still improving. This motivates adjusting $\omega_i(t)$ over the course of training rather than holding it fixed, and we test the idea through an ablation on dynamic weight adaptation. The method we evaluate, CoBa~\citep{coba}, sets each task domain's weight from its absolute and relative evaluation loss, down-weighting early-converging task domains and up-weighting late-converging ones, which is precisely the behavior the second issue calls for. Because the original CoBa experiments were confined to text-modality tasks, we ran these ablations also to test whether the method transfers to the speech-modality tasks of interest. We evaluated it under two task-domain-proportion conditions: an equal-proportion condition, and a redistributed condition with less GR and AR and more Text-IF and Conversational QA domain datasets. The redistributed condition was motivated by our observation that the evaluation losses of accent and gender recognition converge early, freeing budget that can be reallocated to Text-IF and Conversational QA task domains that have not yet converged. This gives four runs in total (CoBa and non-CoBa under each condition), all using 256k training samples under identical training parameters. The results are shown in Figure~\ref{fig:eval-loss}. For AR and GR, the final evaluation loss is comparable across all four runs. For ASR, both non-CoBa runs achieve lower loss than their CoBa counterparts under either condition. For Text-IF and Conversational QA, CoBa under the redistributed condition attains the lowest loss. Since Conversational QA performance is our use-case priority, we adopted the CoBa redistributed strategy for its lowest Conversational QA evaluation loss, accepting the trade-off in ASR.

\subsection{Results}
Adopting the CoBa redistributed strategy mentioned in Section \ref{sec:multi-task-training}, we scaled the training set up to 1 million samples, amounting to an estimated 3{,}000 hours of audio data and 500M text tokens used to adapt the Voxtral-Mini-3B-2507 model. Training was carried out on 2$\times$ NVIDIA H100 NVL GPUs (94\,GB each) with a per-device batch size of 1, gradient accumulation of 8, a single epoch over 62{,}500 steps, a learning rate of $1\times10^{-5}$, and LoRA hyperparameters $r=32$, $\alpha=64$, and dropout $=0.05$. Training completed in 100 hours, yielding our adapted model for the Singaporean context, \textbf{HT-Moonstone}. \\

\begin{table*}[tp]
    \centering
    \caption{Benchmark of candidate SLMs versus HT-Moonstone across the five target tasks. Details of datasets used for each tasks are listed in Appendix \ref{app:first}.
Best result per column in \textbf{bold}. ($\uparrow$/$\downarrow$ indicate that
higher/lower is better.)}
    \label{tab:train-model-benchmark}
    \begin{tabular}{lcccccc}
        \toprule
        & & \textbf{ASR - EN} & \textbf{AR} & \textbf{GR} & \textbf{Spoken QA} & \textbf{Speech QA} \\
        \textbf{Model} & \textbf{Size (B)} & WER\,$\downarrow$ & Acc.\,$\uparrow$ & Acc.\,$\uparrow$ & Acc.\,$\uparrow$ & Acc.\,$\uparrow$ \\
        \midrule
        MERaLiON-2-3B               & 3.5 &     \textbf{5.05} & 44.4 & 87.7 & 25.4 & 68.2 \\
        Voxtral-Mini-3B-2507        & 5 &       7.25 & 24.7 & 15.7 & 42.2 & 76.4 \\
        audio-flamingo-next-hf      & 8 &       7.87 & 0 & 74.6 & 33.0 & 68.7 \\
        SeaLLMs-Audio-7B            & 8 &       114 & 0 & 38.6 & 17.3 & 40.6 \\
        Kimi-Audio                  & 10 &      8.36 & 3.7 & 79.8 & 48.6 & 71.3 \\
        MERaLiON-2-10B              & 10 &      5.46 & 56 & 75.5 & 45.3 & 73.5 \\
        Qwen2.5-Omni                & 11 &      188 & 0 & 18.2 & 36.0 & 79.0 \\
        Voxtral-Small-24B-2507      & 24 &      7.64 & 8.3 & 0.5 & 42.5 & \textbf{80.0} \\
        Qwen3-Omni                  & 35 &      9.85 & 18.6 & 79.1 & \textbf{61.6} & 76.5 \\
        \midrule
        HT-Moonstone                & 5 &       6.59 & \textbf{72.6} & \textbf{93.9} & 52.1 & 75.1 \\
        \bottomrule
    \end{tabular}
\end{table*}

\begin{table*}[tp]
    \centering
    \caption{Benchmark of candidate SLMs versus HT-Moonstone for multilingual ASR. Details of datasets used for each tasks are listed in Appendix \ref{app:first}.
Best result per column in \textbf{bold}. (lower is better.)}
    \label{tab:multilingual-asr}
    \begin{tabular}{lccccccc|c}
        \toprule
        & & \textbf{EN} & \textbf{ZH} & \textbf{ZH-CS} & \textbf{MS} & \textbf{MS-CS} & \textbf{TA} & \textbf{Avg.}\\
        \textbf{Model} & \textbf{Size (B)} & WER\,$\downarrow$ & CER\,$\downarrow$ & WER\,$\downarrow$ & WER\,$\downarrow$ & WER\,$\downarrow$ & CER\,$\downarrow$ & ER\,$\downarrow$ \\
        \midrule
        MERaLiON-2-3B               & 3.5 &       \textbf{5.05} & 7.61 & \textbf{16.4} & \textbf{11.9} & 35.2 & \textbf{8.39} & 14.1 \\
        Voxtral-Mini-3B-2507        & 5 &       7.25 & 18.9 & 60.8 & 41.5 & 100.2 & 85.6 & 52.4 \\
        MERaLiON-2-10B              & 10 &      5.46 & \textbf{6.68} & 19.5 & 12.4 & 28.0 & 8.91 & 13.5 \\
        \midrule
        HT-Moonstone                & 5 &       6.59 & 9.26 & 26.6 & 13.9 & \textbf{20.5} & 10.8 & 14.6 \\
        \bottomrule
    \end{tabular}
\end{table*}

Table~\ref{tab:train-model-benchmark} compares HT-Moonstone against the benchmarked models across the target speech tasks. Most strikingly, HT-Moonstone improves over its base model, Voxtral-Mini-3B-2507, on four of the five tasks: raising accent recognition from 24.7 to 72.6 and gender recognition from 15.7 to 93.9, while sacrificing only roughly one accuracy point on Speech QA, demonstrating effective adaptation while minimising catastrophic forgetting. On accent and gender recognition it attains the best result of any model, surpassing systems up to 7$\times$ its size. On Spoken QA it is second only to Qwen3-Omni, a model 7$\times$ larger, while outperforming all others including its base 24B Voxtral-Small variant. On Speech QA it remains within 5 accuracy points of the best-performing model. We further evaluate HT-Moonstone against SLMs trained from scratch for multilingual ASR across the four languages of Singapore, including Mandarin and Bahasa Melayu code-switching; as shown in Table~\ref{tab:multilingual-asr}, HT-Moonstone reduces its base model's average error rate from 52.4 to 14.6, trailing the purpose-built MERaLiON-2-10B by only one error rate point, and achieving the best result of all on Malay code-switched speech.

\section{Limitations}
We have identified the following limitations in our adaptation methodology:
\begin{itemize}
    \item \textbf{Model size:} Owing to compute constraints, our experiments were confined to a single 5B-parameter SLM (Voxtral-Mini-3B-2507). We were therefore unable to verify whether our findings on catastrophic forgetting, task-aggregated loss, and CoBa-based weighting hold at larger scales, where convergence dynamics and the capacity for multi-task learning may differ.
    \item \textbf{Task coverage:} Our study focuses on a fixed set of speech tasks. We shortlisted additional candidates, such as audio-scene QA and audio emotion recognition, as promising directions for extending the model's capabilities, but were unable to investigate them within the time available. A broader and more diverse task suite would give a more complete picture of how well the approach generalises across speech tasks.
    \item \textbf{Model architecture:} We did not explore swapping out architectural components of the SLM, and whether alternative SLM architectural designs yield better multi-task adaptation remains an open question we hope to pursue.
    \item \textbf{Hyperparameter search:} Our exploration of the training data mixture was necessarily limited. A more exhaustive search over task domain proportions rather than the heuristic reallocation adopted in the redistributed conditions mentioned in Section \ref{sec:multi-task-training} may yield a more optimal mixture and further improve the resultant model.
    \item \textbf{Data imbalance across languages:} As shown in Table~\ref{tab:train-dataset-sizes}, the Tamil and Malay ASR datasets are substantially smaller than those for the other languages. This imbalance prevented us from scaling up the total training set further without over-sampling these low-resource languages. Acquiring additional Tamil and Malay ASR data could further reduce evaluation losses for ASR, Text-IF and Conversational QA task domains and improve general QA performance.
    \item \textbf{Scope of evaluation:} Our evaluation focuses on speech-centric tasks. We did not conduct a comprehensive assessment on broader text-only benchmarks, leaving the extent to which the model's general text-modality capabilities are preserved an open question.
\end{itemize}

\section{Conclusion}
We presented a strategy for adapting an open-source spoken language model to the Singaporean context without access to its original training data. By combining LoRA-based parameter-efficient fine-tuning, a surrogate text-QA grounding dataset to guard against catastrophic forgetting, and a multi-task objective that averages cross-entropy per task domain and adapts the CoBa convergence balancer to speech tasks, we adapted Voxtral-Mini-3B-2507 across five perception and cognitive speech tasks spanning Singapore's four official languages. To enable the spoken-query, text-response setting central to our use cases, we constructed \textbf{HTD-multilingual-QA}, a multilingual QA dataset in both text and spoken form grounded in Singaporean Home Team context. The resulting model, \textbf{HT-Moonstone}, matches or outperforms open-source SLMs up to 7$\times$ its size on most target tasks, attaining the best accent and gender recognition accuracy of any model evaluated, while incurring under 2\% degradation in its original speech QA capability. These results show that targeted, economical adaptation is a viable alternative to training from scratch for specialised, data- and resource-constrained domains. Future work includes validating the approach at larger model scales, broadening the task suite and evaluation benchmarks, and exploring alternative architectural components such as the audio encoder.

\newpage
\section*{Acknowledgments}
The author would like to thank the management and colleagues at Home Team Science \& Technology Agency (HTX) especially Chow Yew Wah, Lim Ming En, Choy Xin Yun Calista, Lye En Lih, Chan Shing Yee and our director Lim Kian Boon, for their support, valuable feedback, and the resources provided throughout this work. Every reasonable effort has been made to acknowledge and cite the original sources of information, ideas, figures, and other materials used in this work. Any inadvertent omissions are unintentional, and the author welcome corrections where appropriate.

\newpage
\bibliography{references}

@article{slm_survey,
  author  = {Siddhant Arora and Kai-Wei Chang and Chung-Ming Chien and Yifan Peng
             and Haibin Wu and Yossi Adi and Emmanuel Dupoux and Hung-Yi Lee
             and Karen Livescu and Shinji Watanabe},
  title   = {On the Landscape of Spoken Language Models: A Comprehensive Survey},
  journal = {arXiv preprint arXiv:2504.08528},
  year    = {2025},
}

@article{slm_tasks,
  author  = {Jing Peng and Yucheng Wang and Bohan Li and Yiwei Guo and Hankun Wang
             and YanGui Fang and Yu Xi and Haoyu Li and Xu Li and Ke Zhang
             and Shuai Wang and Kai Yu},
  title   = {A Survey on Speech Large Language Models for Understanding},
  journal = {arXiv preprint arXiv:2410.18908},
  year    = {2024},
}

@article{kimiaudio,
  author  = {{Kimi Team}},
  title   = {{Kimi-Audio} Technical Report},
  journal = {arXiv preprint arXiv:2504.18425},
  year    = {2025},
}

@article{audioflamingo,
  author  = {Sreyan Ghosh and Arushi Goel and Kaousheik Jayakumar and Lasha Koroshinadze
             and Nishit Anand and Zhifeng Kong and Siddharth Gururani and Sang-gil Lee
             and Jaehyeon Kim and Aya Aljafari and Chao-Han Huck Yang and Sungwon Kim
             and Ramani Duraiswami and Dinesh Manocha and Mohammad Shoeybi
             and Bryan Catanzaro and Ming-Yu Liu and Wei Ping},
  title   = {{Audio Flamingo Next}: Next-Generation Open {Audio}-Language Models
             for Speech, Sound, and Music},
  journal = {arXiv preprint arXiv:2604.10905},
  year    = {2026},
}

@article{meralion,
  author  = {{MERaLiON Team}},
  title   = {{MERaLiON-AudioLLM}: Bridging Audio and Language with Large Language Models},
  journal = {arXiv preprint arXiv:2412.09818},
  year    = {2024},
}

@article{voxtral,
  author  = {{Mistral AI}},
  title   = {{Voxtral}},
  journal = {arXiv preprint arXiv:2507.13264},
  year    = {2025},
}

@article{mnsc,
  author  = {Bin Wang and Xunlong Zou and Shuo Sun and Wenyu Zhang
             and Yingxu He and Zhuohan Liu and Chengwei Wei
             and Nancy F. Chen and AiTi Aw},
  title   = {Advancing {Singlish} Understanding: Bridging the Gap with Datasets
             and Multimodal Models},
  journal = {arXiv preprint arXiv:2501.01034},
  year    = {2025},
}

@article{typhoonaudio,
  author  = {Potsawee Manakul and Guangzhi Sun and Warit Sirichotedumrong
             and Kasima Tharnpipitchai and Kunat Pipatanakul},
  title   = {Enhancing Low-Resource Language and Instruction Following
             Capabilities of Audio Language Models},
  journal = {arXiv preprint arXiv:2409.10999},
  year    = {2024},
}

@article{seallmsaudio,
  author  = {Chaoqun Liu and Mahani Aljunied and Guizhen Chen and Hou Pong Chan
             and Weiwen Xu and Yu Rong and Wenxuan Zhang},
  title   = {{SeaLLMs-Audio}: Large Audio-Language Models for {Southeast Asia}},
  journal = {arXiv preprint arXiv:2511.01670},
  year    = {2025},
}

@article{cantoasr,
  author  = {Dazhong Chen and Yi-Cheng Lin and Yuchen Huang and Ziwei Gong
             and Di Jiang and Zeying Xie and Yi R. Fung},
  title   = {{CantoASR}: Prosody-Aware {ASR-LALM} Collaboration for Low-Resource
             {Cantonese}},
  journal = {arXiv preprint arXiv:2511.04139},
  year    = {2025},
}

@article{slu_limited_data,
  author  = {Youngwon Choi and Jaeyoon Jung and Hyeonyu Kim
             and Huu-Kim Nguyen and Hwayeon Kim},
  title   = {Exploring Fine-Tuning of Large Audio Language Models for Spoken
             Language Understanding under Limited Speech Data},
  journal = {arXiv preprint arXiv:2509.15389},
  year    = {2025},
}

@article{dpo_codeswitch,
  author  = {Trung Nguyen Quang and Cheng Yi Lewis Won and Minh Duc Pham
             and Yingxu He and Shuo Sun and Ai Ti Aw},
  title   = {Direct Preference Optimization for {English}-{Mandarin} Code-Switching
             Speech Recognition in Audio {LLMs}},
  journal = {arXiv preprint arXiv:2605.23975},
  year    = {2026},
}

@article{lora,
  author  = {Edward J. Hu and Yelong Shen and Phillip Wallis and Zeyuan Allen-Zhu
             and Yuanzhi Li and Shean Wang and Lu Wang and Weizhu Chen},
  title   = {{LoRA}: Low-Rank Adaptation of Large Language Models},
  journal = {arXiv preprint arXiv:2106.09685},
  year    = {2021},
}

@article{omnivoice,
  author  = {Han Zhu and Lingxuan Ye and Wei Kang and Zengwei Yao and Liyong Guo
             and Fangjun Kuang and Zhifeng Han and Weiji Zhuang and Long Lin
             and Daniel Povey},
  title   = {{OmniVoice}: Towards Omnilingual Zero-Shot Text-to-Speech with
             Diffusion Language Models},
  journal = {arXiv preprint arXiv:2604.00688},
  year    = {2026},
}

@article{coba,
  author  = {Zi Gong and Hang Yu and Cong Liao and Bingchang Liu
             and Chaoyu Chen and Jianguo Li},
  title   = {{CoBa}: Convergence Balancer for Multitask Finetuning of Large
             Language Models},
  journal = {arXiv preprint arXiv:2410.06741},
  year    = {2024},
}

@article{whisper,
  author  = {Alec Radford and Jong Wook Kim and Tao Xu and Greg Brockman
             and Christine McLeavey and Ilya Sutskever},
  title   = {Robust Speech Recognition via Large-Scale Weak Supervision},
  journal = {arXiv preprint arXiv:2212.04356},
  year    = {2022},
}

@inproceedings{squadsrc,
  author    = {Yixuan Tang and Anthony K. H. Tung},
  title     = {{SQuAD-SRC}: A Dataset for Multi-Accent Spoken Reading Comprehension},
  booktitle = {Proceedings of the Thirty-Second International Joint Conference
               on Artificial Intelligence (IJCAI-23)},
  pages     = {5206--5214},
  year      = {2023},
}

@inproceedings{audiobench,
  author    = {Bin Wang and Xunlong Zou and Geyu Lin and Shuo Sun and Zhuohan Liu
               and Wenyu Zhang and Zhengyuan Liu and AiTi Aw and Nancy F. Chen},
  title     = {{AudioBench}: A Universal Benchmark for Audio Large Language Models},
  booktitle = {Proceedings of the 2025 Conference of the Nations of the Americas
               Chapter of the Association for Computational Linguistics (NAACL)},
  year      = {2025},
}


\newpage
\appendix
\section{List of Benchmarking Datasets}
\label{app:first}
The below are the list of benchmarking datasets used, arranged in tasks mentioned:

\begin{sloppypar}
\begin{enumerate}
  \item \textbf{ASR-EN:} \path{MNSC/ASR-Part1-Test}
  \item \textbf{ASR-ZH:} \path{aishell1_mandarin_test}, \path{aishell3_mandarin_test}, \path{fleurs_mandarin_test}
  \item \textbf{ASR-ZH-CS:} \path{seame_dev_man}, \path{seame_dev_sge}
  \item \textbf{ASR-MS:} \path{fleurs_malay_test}
  \item \textbf{ASR-MS-CS:} \path{mesolitica_malay_test}
  \item \textbf{ASR-TA:} \path{fleurs_tamil_test}, \path{slr65_tamil_test}, \path{slr127_tamil_test}
  \item \textbf{AR:} \path{MNSC/PQA-AR-Dialogue-Test}
  \item \textbf{GR:} \path{MNSC/PQA-GR-Dialogue-Test}
  \item \textbf{Spoken QA:} internally developed, covering Home Team context across the four Singaporean languages
  \item \textbf{Speech QA:} \path{AudioLLMs/public_sg_speech_qa_test}
\end{enumerate}
\end{sloppypar}

\end{document}